\RequirePackage{fix-cm}
\documentclass[smallextended]{svjour3}       \smartqed  \usepackage{graphicx}
\usepackage[T1]{fontenc}
\usepackage[misc]{ifsym}

\usepackage{hyperref}

\usepackage{subcaption}
\usepackage{multirow}

\newcommand*{\affaddr}[1]{#1} 
\newcommand*{\affmark}[1][*]{\textsuperscript{#1}}

\newcommand{\modelname}{AD-AutoGPT} 
\begin{document}

\title{
\modelname: An Autonomous GPT for Alzheimer's Disease Infodemiology
}


\author{Haixing Dai\protect\affmark[1], Yiwei Li\protect\affmark[1], Zhengliang Liu\protect\affmark[1], Lin Zhao\protect\affmark[1], Zihao Wu\protect\affmark[1], Suhang Song\protect\affmark[2], Ye Shen\protect\affmark[3], Dajiang Zhu\protect\affmark[4], Xiang Li\protect\affmark[5,6], Sheng Li\protect\affmark[7], Xiaobai Yao\protect\affmark[8], Lu Shi\protect\affmark[9], Quanzheng Li\protect\affmark[5,6], Zhuo Chen\protect\affmark[2], Donglan Zhang\protect\affmark[10], Gengchen Mai\protect\affmark[8,1*], Tianming Liu\protect\affmark[1**]          }

\authorrunning{Dai et al.}

\institute{
 Haixing Dai \\   \email{hd54134@uga.edu}\\\\
 Yiwei Li \\   \email{yl80817@uga.edu}\\\\
 Zhengliang Liu \\   \email{zl18864@uga.edu}\\\\
 Lin Zhao \\   \email{lin.zhao@uga.edu}\\\\
 Zihao Wu \\   \email{zw63397@uga.edu}\\\\
 Suhang Song \\   \email{suhang.song@uga.edu}\\\\
 Ye Shen \\   \email{yeshen@uga.edu}\\\\
 Dajiang Zhu \\   \email{dajiang.zhu@uta.edu}\\\\
 Xiang Li \\   \email{xli60@mgh.harvard.edu}\\\\
 Sheng Li \\   \email{lisheng1989@gmail.com}\\\\
 Xiaobai Yao \\   \email{xyao@uga.edu}\\\\
 Lu Shi \\   \email{LUS@clemson.edu}\\\\
 Quanzheng Li \\   \email{li.quanzheng@mgh.harvard.edu}\\\\
 Zhuo Chen \\   \email{zchen1@uga.edu}\\\\
 Donglan Zhang \\   \email{Donglan.Zhang@nyulangone.org}\\\\
 \Letter $\;\;$ Gengcheng Mai \\   \email{gengchen.mai25@uga.edu}\\\\
 \Letter $\;\;$ Tianming Liu \\   \email{tliu@uga.edu}\\\\
 \affaddr{\affmark[1]School of Computing, University of Georgia.} \\
 \affaddr{\affmark[2]College of Public Health, University of Georgia.} \\
 \affaddr{\affmark[3]Department of Epidemiology and Biostatistics, University of Georgia.} \\
 \affaddr{\affmark[4]Department of Computer Science and Engineering, University of Texas at Arlington.} \\
 \affaddr{\affmark[5]Massachusetts General Hospital.} \\
 \affaddr{\affmark[6]Harvard Medical School.} \\
 \affaddr{\affmark[7]School of Data Science, University of Virginia.} \\
 \affaddr{\affmark[8]Department of Geography, University of Georgia.} \\
 \affaddr{\affmark[9]Department of Public Health Science, Clemson University.} \\
 \affaddr{\affmark[10]NYU Long Island School of Medicine, New York University.} \\
 \affaddr{\affmark[*]Corresponding author.} \\
 \affaddr{\affmark[**]Senior Corresponding author.} \\
 }

\date{Received: date / Accepted: date}

\maketitle

\begin{abstract}
In this pioneering study, inspired by AutoGPT, the state-of-the-art open-source application based on the GPT-4 large language model, we develop a novel tool called \modelname{} which can conduct data collection, processing, and analysis about complex health narratives of Alzheimer's Disease in an autonomous manner via users' textual prompts. 
We collated comprehensive data from a variety of news sources, including the Alzheimer's Association, BBC, Mayo Clinic, and the National Institute on Aging since June 2022, leading to the autonomous execution of robust trend analyses, intertopic distance maps visualization, and identification of salient terms pertinent to Alzheimer's Disease. This approach has yielded not only a quantifiable metric of relevant discourse but also valuable insights into public focus on Alzheimer's Disease. This application of \modelname{} in public health signifies the transformative potential of AI in facilitating a data-rich understanding of complex health narratives like Alzheimer's Disease in an autonomous manner, setting the groundwork for future AI-driven investigations in global health landscapes.
\subclass{MSC code1 \and MSC code2 \and more}
\end{abstract} \section{Introduction}
\label{intro}
Alzheimer's Disease (AD), a progressive neurodegenerative disorder, remains one of the most pressing public health concerns globally in the 21st century \cite{avramopoulos2009genetics,dartigues2009alzheimer}. This disease, characterized by cognitive impairments such as memory loss, predominantly affects aging populations, exerting an escalating burden on global healthcare systems as societies continue to age \cite{post2000moral}. The significance of AD is further magnified by the increasing life expectancy globally, with the disease now recognized as a leading cause of disability and dependency among older people \cite{hinton1999constructing}. Consequently, AD has substantial social, economic, and health system implications, making its understanding and awareness of paramount importance \cite{rice1993economic,zhao2008healthcare}.

Despite the ubiquity and severity of AD, a gap persists in comprehensive, data-driven public understanding of this complex health narrative. Traditionally, public health professionals have to rely on labor-intensive methods such as web scraping, API data collection, data postprocessing, and analysis/synthesis to gather insights from news media, health reports, and other textual sources \cite{zhang2021monitoring,bacsu2022using,mavragani2020infodemiology}. However, these methods often necessitate complex pipelines for data gathering, processing, and analysis. Moreover, the sheer scale of global data presents an ever-increasing challenge, one that demands a novel, innovative approach to streamline these processes and extract valuable, actionable insights efficiently and automatically. In addition, the technical expertise required for developing data processing and analysis pipelines significantly limits the access and engagement of the broader public health community. 

AutoGPT \cite{richards2023auto} is an experimental open-source application that harnesses the capabilities of large language models (LLMs) such as GPT-4 \cite{openai2023gpt} and ChatGPT \cite{liu2023summary} to automate and optimize the analytical process. With its advanced linguistic understanding and autonomous operation, AutoGPT simplifies complex data pipelines, facilitating comprehensive analyses of vast datasets with simple textual prompts. This tool transcends traditional limitations, unlocking the potential of LLMs for autonomous data collection, processing, summarization, analysis, and synthesis. 
In this study, we modify the AutoGPT architecture into public health applications and develop \modelname{} to analyze a multitude of news sources, including the Alzheimer's Association, BBC, Mayo Clinic, and the National Institute on Aging, focusing on discourse since June 2022. We are among the pioneers in integrating AutoGPT into public health informatics, adapting this transformative AI tool into the public health domain to elucidate the complex narrative surrounding Alzheimer's Disease. This research underlines the enormous potential of autonomous LLMs in global health research, paving the way for future AI-assisted investigations into various health-related domains.

We summarize our key contributions below:
\begin{itemize}
    \item Inspired by AutoGPT, we develop a novel LLM-based tool called \modelname{}, which can generate data collection, processing, and analysis pipeline in an autonomous manner based on users' textual prompts. More specifically, we adapt \modelname{} to the public health domain to showcase its great potential of autonomous pipeline generation to understand the complex narrative surrounding Alzheimer's Disease. 

    \item While AutoGPT is an effective autonomous LLM-based tool, it has lots of limitations when applying it on AD Infodemiology during the process of public health information retrieval, text-based information extraction, text summarization, summary analysis, and visualization.
To overcome AutoGPT's limitations for the AD Infodemiology task, \modelname{} provides the following improvements: 1) specific prompting mechanisms to improve the efficiency and accuracy of AD information retrieval; 
    2) a tailored spatiotemporal information extraction functionality; 
    3) an improved text summarization ability;
    4) an in-depth analysis ability on generated text summaries;
    and 5) an effective and dynamic visualization capability.
    
    \item We show that \modelname{}
    transforms the traditional labor-intensive data collection, processing, and analysis paradigm into a prompt-based automated, and optimized analytical framework. This has allowed for efficient, comprehensive analysis of numerous news sources related to Alzheimer's Disease.
    
    \item Through \modelname{}, we have provided a case study for detailed trend analysis, intertopic distance mapping, and identified salient terms related to Alzheimer's Disease from four AD-related new sources. This contributes significantly to the existing body of knowledge and facilitates a nuanced understanding of the disease's discourse in public health.
    
    \item Our research underlines the capacity of \modelname{} to facilitate data-driven public understanding of complex health narratives, such as Alzheimer's Disease, which is of paramount importance in an aging global society.
    
    \item The methodologies and insights from our work provide a foundation for future AI-assisted public health research. Our \modelname{} pipeline is extendable to other topics in public health or even other domains. This work paves the way for comprehensive and efficient investigations into various domains.
\end{itemize} \section{Related Work}
\label{re}
\subsection{Large Language Models}
\label{re:1}
Large language models (LLMs), with their origins in Transformer-based pre-trained language models (PLMs) such as BERT \cite{devlin2018bert} and GPT \cite{radford2018improving}, have substantially transformed the field of natural language processing (NLP). LLMs have superseded previous methods such as Recurrent Neural Network (RNN) based models, leading to their widespread adoption across various NLP tasks \cite{liu2023summary,zhao2023brain}. Furthermore, the emergence of very large language models such as GPT-3 \cite{brown2020language}, Bloom \cite{scao2022bloom}, GPT-4 \cite{openai2023gpt}, PaLM \cite{chowdhery2022palm}, and PaLM-2 \cite{anil2023palm} demonstrates a clear trend towards even more sophisticated language understanding capabilities.

These models are designed to learn accurate contextual latent feature representations from input text \cite{kalyan2021ammus}, which can then be employed in a variety of applications, including question answering, information extraction, sentiment analysis, text classification, and text generation. The innovative technique of reinforcement learning from human feedback (RLHF) \cite{ziegler2019fine} has been used to further align LLMs with human preferences, which has found applications in Artificial General Intelligence (AGI) models such as InstructGPT \cite{ouyang2022training}, Sparrow \cite{glaese2022improving}, and ChatGPT \cite{liu2023summary}. More recently, GPT-4 has significantly advanced the state-of-the-art of language models, opening up new opportunities for LLM applications.

Other than the applications in NLP domain, LLMs also show promising results and significant impacts in other disciplines such as biology \cite{agathokleous2023use}, geography \cite{mai2022towards,mai2023opportunities}, agriculture \cite{lu2023agi}, education \cite{kasneci2023chatgpt,latif2023artificial}, medical and health care \cite{liu2023deid,dave2023chatgpt}, and so on.

\subsection{Public Health Infodemiology}
\label{re:2}
Infodemiology \cite{eysenbach2002infodemiology} is a field that studies the determinants and distribution of information on the internet or in a population, with the goal of informing public health and public policy \cite{eysenbach2002infodemiology,mavragani2020infodemiology}. The term combines "information" and "epidemiology" and is a recognized approach in public health informatics, providing insights into health-related behaviors and perceptions. It plays a crucial role in monitoring and managing the information epidemic ("infodemic") associated with major public health crises.

For example, Piamonte et al. \cite{piamonte2022googling} analyzed global search queries for Alzheimer's disease (AD) using Google Trends data, comparing this online interest (Search Volume Index) with measures of disease burden. The study revealed that search behavior and interest in AD were influenced by factors like news about celebrities with AD and awareness months, and also highlighted potential correlations between this online interest and socioeconomic development.

With the rise of the internet and digital technologies, infodemiology provides a vital lens to examine the flow of health information and misinformation, helping public health practitioners develop effective communication strategies and interventions \cite{zielinski2021infodemics,mackey2022advancing}. In the context of Alzheimer's disease, understanding online behaviors and interests via infodemiology can help enhance public awareness, correct misconceptions, and inform preventative and management strategies for the disease \cite{piamonte2022googling,bacsu2022examining}.

\subsection{AutoGPT and LLM Automation}
\label{re:3} 
The development and use of AutoGPT, LangChain\footnote{\url{https://python.langchain.com/en/latest/index.html}}, and many other automation techniques for LLMs represent a significant advancement in the field of NLP and artificial intelligence. AutoGPT builds on the successes of large language models like GPT-3 and GPT-4, but takes automation a step further by providing a more user-friendly interface for non-expert users \cite{richards2023auto}. 

With AutoGPT, complex tasks such as data collection, data cleaning, analysis, and even the generation of human-like text can be completed using straightforward prompts, removing the need for extensive coding or data science expertise. This has the potential to democratize access to powerful language model technology, opening up new possibilities for research and application in a wide range of fields, including public health.

Recent studies \cite{zhaocomprehensive,fezarigpt} have highlighted the potential of AutoGPT and similar tools for automating the retrieval and analysis of large datasets. For example, with a well-formulated query, AutoGPT can be directed to crawl through a wide array of online platforms, collecting and analyzing comments, discussions, and posts pertaining to vaccines. The system would subsequently generate a summarizing report, outlining major themes of public opinion and prevalent misconceptions, thereby providing valuable insights for public health officials in formulating targeted communication and intervention strategies.

In the context of infodemiology, AutoGPT can automate the process of analyzing online health information trends, which traditionally involves extensive manual effort. Specifically, it can efficiently scan and interpret internet data, track the spread of health information and misinformation, assess public reaction to health policies or events, and potentially predict future trends.

\subsection{Improving Autonomous LLM-based Tools for Public Health}

While recognizing the potential of autonomous large language models (LLMs) like AutoGPT in public health research and practice, we identified certain limitations in their current state that may hinder their efficacy in particular use cases, such as infodemiology. By tailoring these tools to the specific needs of public health professionals, we aim to enhance their utility in these contexts.

Firstly, despite AutoGPT's extensive searching capabilities, its ability to acquire specialized information quickly and precisely, for instance, about Alzheimer's disease (AD), can be somewhat limited. In response to this, we have integrated specific prompting mechanisms in our model, \modelname{}. These tailored prompts direct \modelname{} to gather data from a select list of authoritative websites relevant to AD, which enhances the efficiency and relevance of information acquisition.

Secondly, Our \modelname{} model also addresses the challenge AutoGPT faces in extracting critical details such as the time and place of news events from articles accurately. \modelname{} uses web-crawling scripts to extract accurate timestamps from news pieces, and employs geo-location libraries such as geopy \cite{githubGitHubGeopygeopy} and geopandas \cite{githubGitHubGeopandasgeopandas} to retrieve precise location information from texts.

Thirdly, depth of analysis is another area where AutoGPT could benefit from further refinement. Owing to the token limit in models like ChatGPT, AutoGPT's analysis is often restricted to the first 4096 tokens \cite{liu2023summary}. Consequently, it might miss core content or important details. To overcome this limitation, \modelname{} segments the text, vectorizes it, and then processes these chunks independently. It creates summaries for each of these segments and then amalgamates these summaries to create a comprehensive representation of the news article.

Fourthly, AutoGPT's current capabilities, while useful, lack the capacity to conduct an in-depth analysis of the generated summaries. The synthesized data can still be redundant and may not accurately capture the most essential information. In contrast, \modelname{} applies Latent Dirichlet Allocation (LDA) \cite{blei2003latent} to extract the most pertinent keywords from the text summaries, offering users a succinct understanding of the central themes in the Alzheimer's disease domain.

Lastly, while AutoGPT is effective at generating text-based information, it lacks robust visualization capabilities. Addressing this limitation, \modelname{} integrates dynamic visualization techniques, creating plots of news occurrences over time, highlighting locations where news events are happening, and even illustrating the evolution of research keywords over time.

\modelname{} is refined through the application of domain-specific knowledge and technical adjustments to optimize its relevance and effectiveness for public health researchers and practitioners. As a result, \modelname{} is faster and more efficient in its operations compared to the original AutoGPT, highlighting the advantages of tailoring autonomous LLM-based tools for specific use cases in public health.

\begin{figure}
\includegraphics[width=13cm]{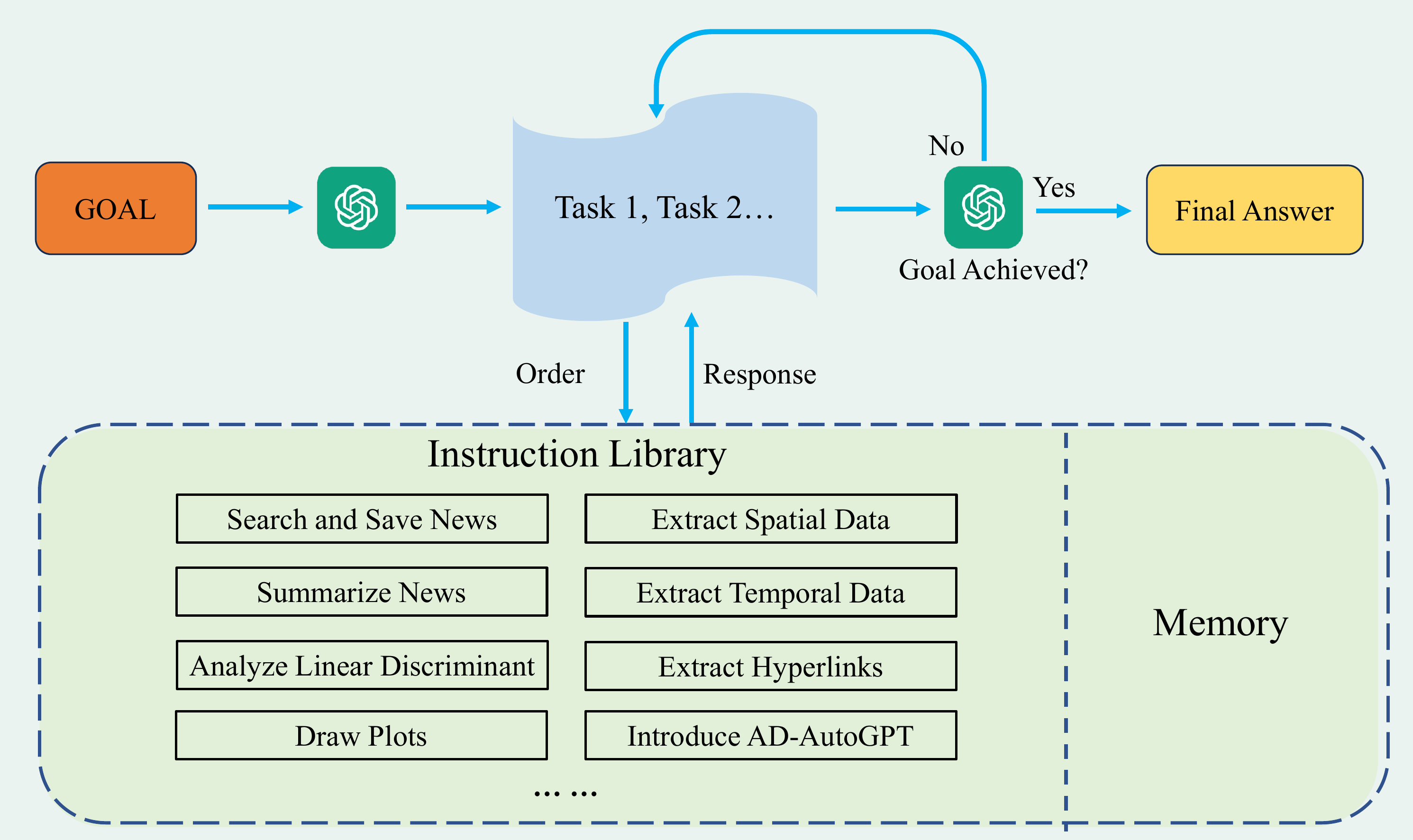}
\caption{The basic framework of \modelname{}. The instruction library contains a set of possible commands we have developed to complete the public health infodemiology task. These commands can also be expanded in the future. 
In order to achieve the goal, \modelname{} will access GPT-4 and divide the final goal into several smaller tasks, and then solve small tasks step-by-step by choosing the most appropriate command for the sub-task in the instruction library. After thinking and judging, if the final goal has not been achieved, \modelname{} will continue to split the task and execute the command. If the final goal has been achieved, \modelname{} will return the final answer.}
\label{fig:framework}       \end{figure}
\section{Method}
\label{sec:method} 
\begin{figure}[ht!]
	\centering \tiny
\begin{subfigure}[b]{1.0\textwidth}  
		\centering 
		\includegraphics[width=\textwidth]{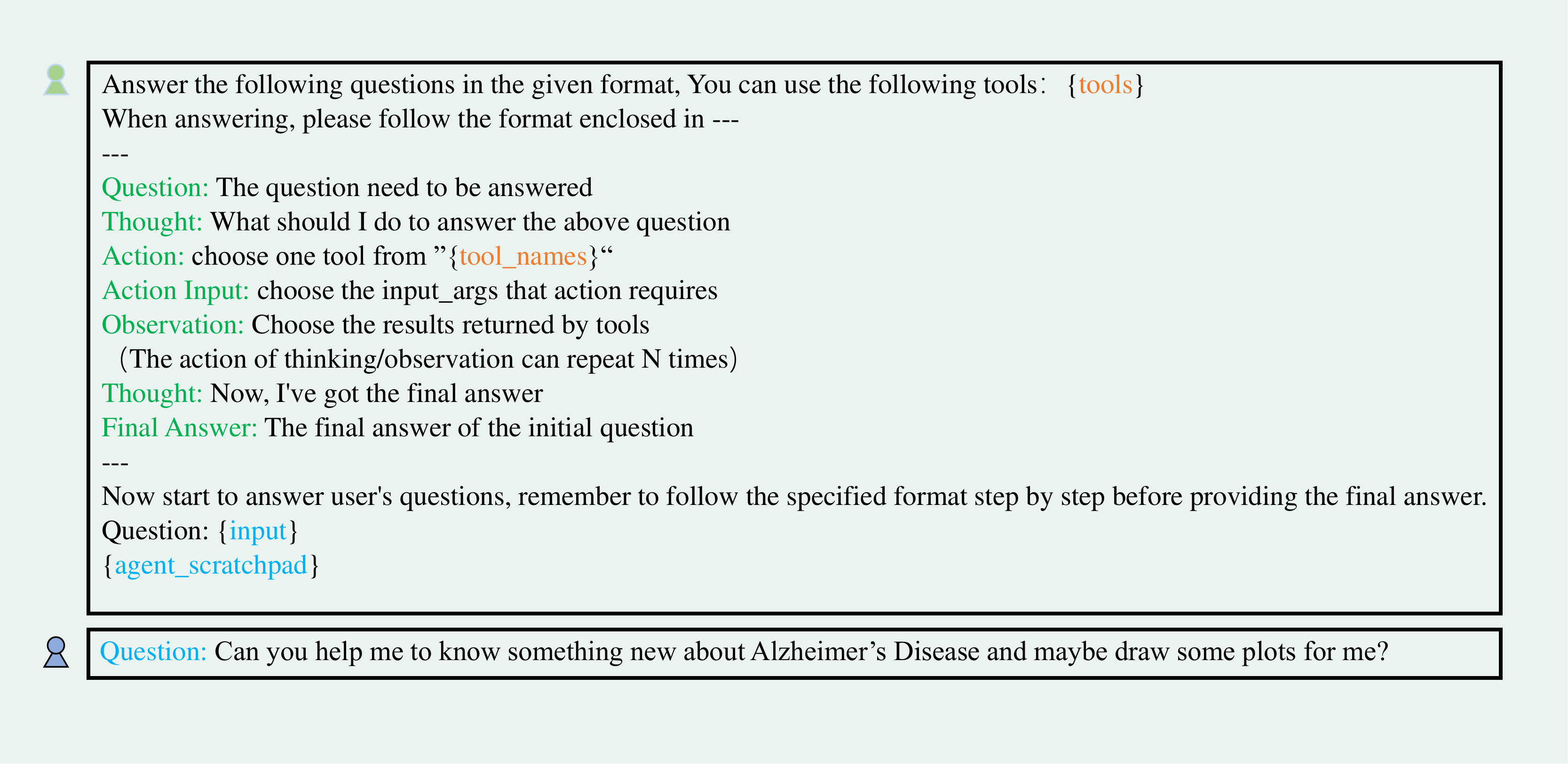}\caption[]{{An instance of prompt specifies the format in which the AI answers questions.
		}}    
		\label{fig:prompt}
	\end{subfigure}
	\hfill
	\begin{subfigure}[b]{1.0\textwidth}  
		\centering 
		\includegraphics[width=\textwidth]{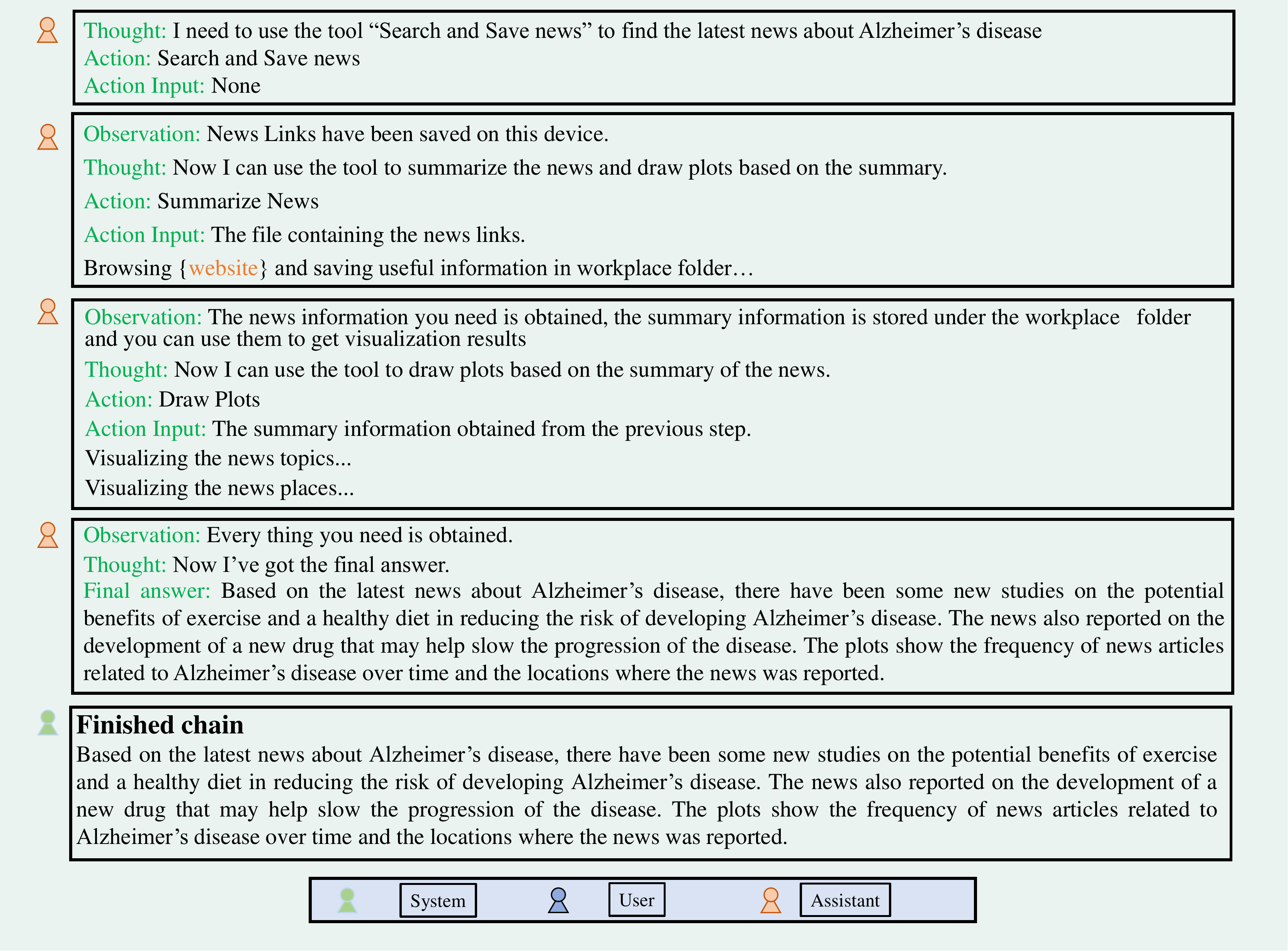}\caption[]{{
An example of AI thinking and calling functions to solve user problems
		}}    
		\label{fig:process_prompt}
	\end{subfigure}
	\hfill
	\caption{{
	The prompt of \modelname{}, the AI assistant will answer the question based on the given format and can use the specified functions. In the
prompt, tools represent the functions that \modelname{} can call, including
tool\_names, tool\_descriptions and so on.}
	} 
	\label{prompt_all}
\end{figure}

In this section, we will introduce \modelname{}, an LLM-based tool we developed to automate the process of Alzheimer's Disease Infodemiology.  \modelname{} uses the Langchain framework to realize the connection with GPT-4 and ChatGPT API, and establish an LLM-based autonomous framework with a chain of thinking mode for Alzheimer's disease. This is a model that can automatically search for the latest news, extract meaningful spatio-temporal data, summarize the news, analysis news content, and visualize analysis results. The overall framework of \modelname{} is shown in Figure \ref{fig:framework}. 
We construct an instruction library that contains a set of possible commands/tools we have developed to achieve the public health infodemiology task. 
A prompt shown in Figure \ref{fig:prompt} is designed to facilitate LLMs to identify usable tools from the instruction library and form a data processing pipeline that demonstrates the process of thinking. \modelname{}'s ability of ``translating'' natural language prompts to real data processing pipeline is similar to the idea of semantic parsing used in traditional question answering literature \cite{liang2017neural,berant2013semantic,mai2021geographic}, which aims at translating a natural language question into an executable query for a given database or knowledge base. 
However, the difference is that semantic parsing is only able to generate rather simple executable queries on a well-defined knowledge base while our \modelname{} can handle much more complex real-world tasks such as searching and collecting news from Google, analyzing new contents, and visualizing topic trends and spatial-temporal distributions of news. 
Below we will introduce the workflow of \modelname{} and the basic principles of the algorithms used in the workflow in detail.

\subsection{Overall Framework}
\label{sec:framework}
Our primary goal is to learn from the chain thinking mode of AutoGPT to realize the automatic collection and summary of Alzheimer's disease news. To achieve this goal, the power of LLMs must be used. Advanced LLMs such as ChatGPT and GPT-4 have brought earth-shaking changes to the NLP domain, and we see the potential advantages of LLMs for the public health field.

The overall framework is shown in Figure \ref{fig:framework}. For the target task, \modelname{} will use ChatGPT or GPT-4 to divide the target task into several small tasks and process them separately. We provide \modelname{} with an instruction library which contains customized functions/tools including: 
\begin{enumerate}
    \item \textbf{Search and Save News}, which utilizes Google API to search for the latest news posted on authoritative websites and save the URLs on a local device;
    \item \textbf{Summarize News}, which uses ChatGPT or GPT-4 to summarize the main content of one piece of news and extract the spatial-temporal information of each stored news;
    \item \textbf{Visualize Results}, which will draw all the results for visualization, and will also display the results of the LDA analysis of the news content. 
\end{enumerate}

After operating every small task choosing from these tools, \modelname{} will judge whether the overall goal has been achieved according to the running results of the function, or it needs to think again and solve the next small problem. Chain thinking is realized through such a pattern. If during the process \modelname{} thinks that the system has reached the initial goal, the system will exit and return a final answer to the initial question.

\subsection{Designing Prompts to Implement Chain of Thoughts}
\label{sec:Prompts}
A prompt example can be seen in Figure \ref{fig:prompt} and the model thinking process of \modelname{} is shown in Figure \ref{fig:process_prompt}. According to the input, this prompt has four parts in the task process which are question, thought, action, and action input. 
\begin{enumerate}
    \item \textbf{Question} is the problem that AI needs to solve. 
    \item \textbf{Thought} is the idea and thought process of AI for this problem.
    \item \textbf{Action} is the operation selected by AI after thinking which AI thinks is most suitable for solving the current task.
    \item \textbf{Action input} is used as the input of the function.
\end{enumerate}
For output, a prompt has three parts which are observation, thought, and final answer. 
\begin{enumerate}
    \item \textbf{Observation} is the output of the function to inspire AI's next thinking.
    \item \textbf{Thought} shows the results of AI's thinking about Observation.
    \item The \textbf{final answer} is the judgment of the result. If the AI thinks that the current result can fully answer the initial question, the AI will return the final answer. Otherwise, it will continue to think and call other functions. 
\end{enumerate}

The last part of the prompt is the question entered by the user, such as the question in Figure \ref{fig:prompt}, "\textit{Can you help me to know something new about Alzheimer’s Disease and maybe draw some plots for me?}". AI will decompose the complex target tasks proposed by users into several simple tasks, thus inspiring a chain of thoughts. And the thinking process of AI can be seen in Figure \ref{fig:process_prompt}

Owing to this set of prompts, we can ensure that the thinking logic of \modelname{} does not deviate from the right track and make the whole chain of thoughts visible to users.

\subsection{Text Summary}
\label{sec:summary} 
To achieve the purpose of extracting the most critical information from a large amount of news text, \modelname{} performs new text summary and LDA topic modeling.

The text summary is mainly achieved by accessing ChatGPT or GPT-4 API. Owing to the powerful text summarization ability of GPT-4, \modelname{} can make more efficient use of text than other models. \modelname{} traverses the saved news URLs one by one, and then saves the text from the website by calling the web crawler scripts. Next, it uses ChatGPT or GPT-4 to summarize the news text. It is worth mentioning that because LLMs have a token limit, all the text here will be pre-processed first, and then be summarized. More specifically, since GPT-4 has a limit on the number of tokens, in order to summarize the complete news text, we use the map\_reduce method to process it \cite{richards2023auto}.

\subsection{Spatiotemporal Information Extraction} \label{sec:spatem_ext}

Next, \modelname{} will perform spatiotemporal information extraction on the collected news articles.  
The temporal information can be easily extracted from news metadata while extracting place mentions from news articles is a kind of oral. Here, we adopt the geoparsing approach \cite{gritta2018melbourne,karimzadeh2019geotxt} which first recognizes place names from raw text, so-call toponym recognition \cite{wang2020neurotpr,mai2022towards} and then link the recognized place names to a specific geographic entity in an existing gazetteer or geospatial knowledge graphs \cite{ahlers2013assessment,mai2022symbolic}, so-called toponym resolution \cite{ju2016things}, so that the spatial footprints (i.e., geographic coordinates) of these places can be obtained.
More specifically, we use GeoText\footnote{\url{https://github.com/elyase/geotext}}, a python-based geoparsing tool to achieve this goal. 

\subsection{LDA Analysis}
\label{sec:lda} 
Latent Dirichlet Allocation (LDA) \cite{blei2003latent} is a probabilistic topic model. LDA can give a probability distribution of topics of each document in the corpus. By analyzing a batch of document sets and extracting their topic distributions, topic clustering can be performed according to the topic distribution. LDA is a typical bag-of-words model, that is, a document is interpreted as a set of words, and there is no sequential relationship among words. In addition, a document can contain multiple topics, and each word in the document is assumed to be generated by one of the topics. LDA is an unsupervised learning method that does not require a manually labeled training set during training but only needs a document set and the total number of topics $K$. In addition, another advantage of LDA is that every topic is associated with a set of most frequent keywords which can be used to interpret this topic. 

In short, \modelname{} uses LDA topic modeling to discover the topics for the
summary text of each piece of collected news, For each topic, the keyword with the highest frequency of occurrence and the highest weight will be displayed to the user. %
 
\section{Case Study and Experimental Results} \label{sec:exp}

\subsection{Alzheimer’s Disease News Information Retrieval}
\label{sec:search} 
The effectiveness of our proposed \modelname{} is mainly verified on the data provided by the most authoritative websites reporting Alzheimer's disease, which are Alzheimer’s Association, BBC, National Institute of Aging, and Mayo Clinic. 
By using the prompt shown in Figure \ref{fig:prompt}, we are able to instruct the LLM (e.g., ChatGPT or GPT-4) to search for the right tool in our instruction library -- \textit{Search and Save News} to achieve the first news data collection step.

We have collected 277 news in total from these four websites in the period of last year. On this actual news dataset, we validate the functions of \modelname{} for text extraction, text summarization, spatio-temporal-data analysis, hot topics analysis, and result visualization. In this process, the time and location of the news will also be extracted and saved. Note that \modelname{} automatically uses the given prompt and formalizes a data collection and processing pipeline based on the toolsets in our instruction library without any human intervention.

\subsection{Spatiotemporal Information Extraction and Visualization} \label{sec:sp-ana}

\begin{figure}[ht!]
	\centering \tiny
\begin{subfigure}[b]{0.48\textwidth}  
		\centering 
		\includegraphics[width=\textwidth]{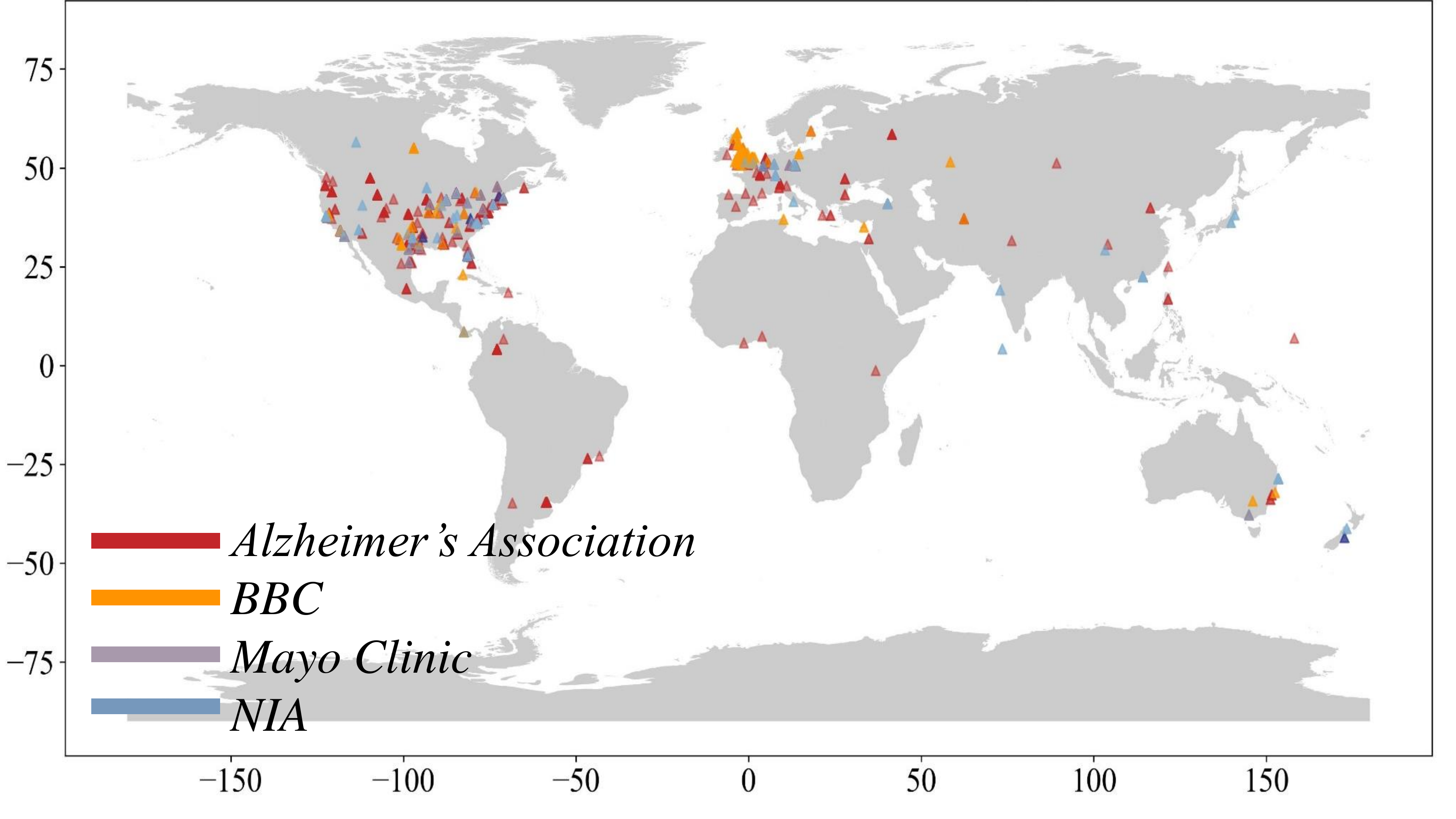}\caption[]{{Places where the latest news about Alzheimer's diseases happened.
		}}    
		\label{fig:Spatial-data}
	\end{subfigure}
	\hfill
	\begin{subfigure}[b]{0.48\textwidth}  
		\centering 
		\includegraphics[width=\textwidth]{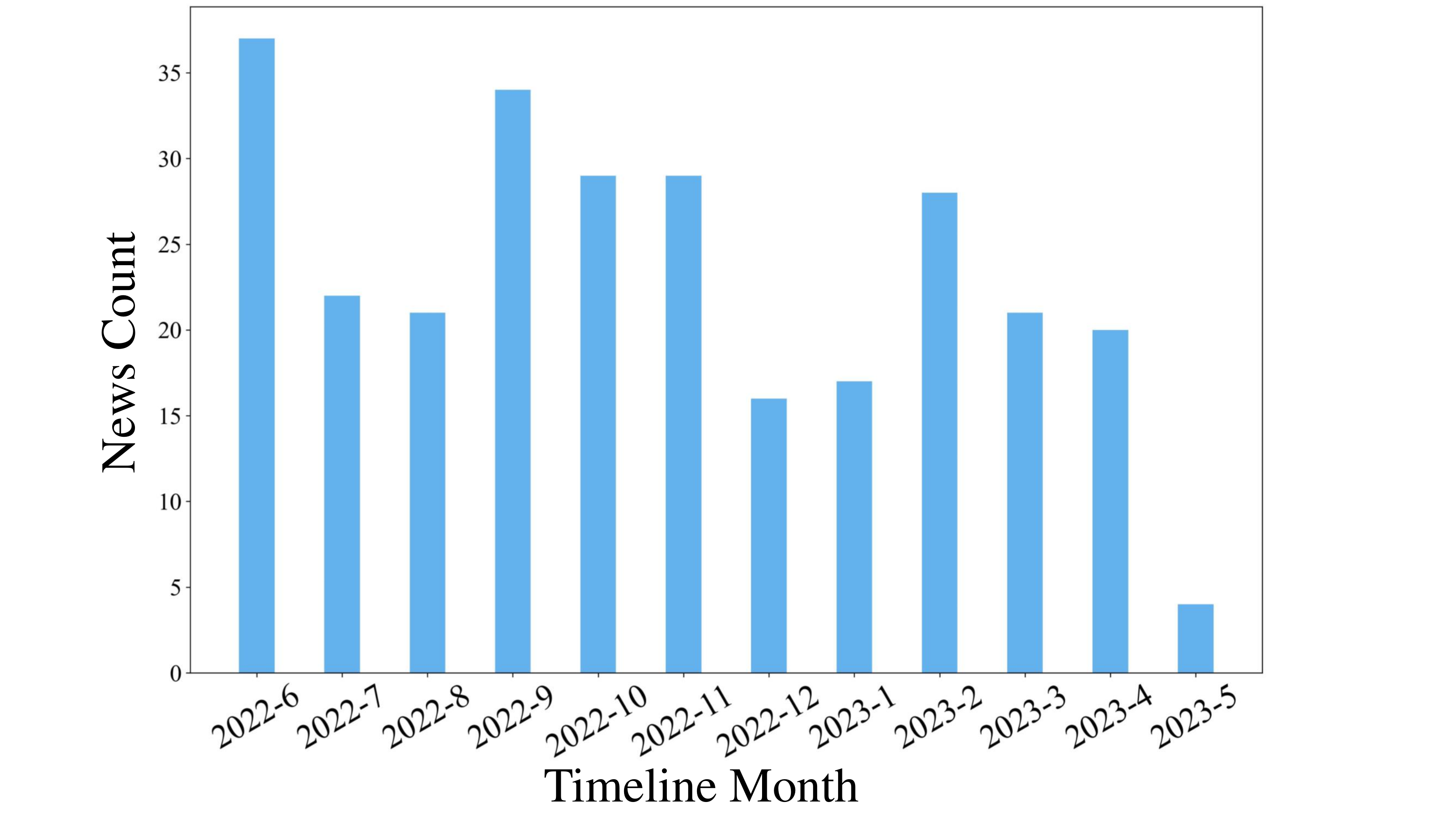}\caption[]{{The number of news collected for each month from June 2022 to May 2023.
		}}    
		\label{fig:temporal}
	\end{subfigure}
	\hfill
	\caption{{
	The visualization of the results from the spatial and temporal information extraction.
	(a) shows the spatial distribution of the Alzheimer's disease news. The news mainly happened in America and 
Western Europe. (b) shows the temporal change in the number of news occurrences from June 2022 to May 2023.}
	} 
	\label{Spatial_and_Temporal}
\end{figure} 
Based on the given prompt, \modelname{} decides to use \textit{Extract Spatial Data} tool and \textit{Extract Temporal Data} tool in our instruction library (see Figure \ref{fig:framework}) to extract the places where these news articles mentioned and the timestamps when these news articles were posted online.

The spatial locations of extracted places from all news articles are visualized in Figure \ref{fig:Spatial-data}. Note that this map visualization is automatically generated by \modelname{} based on the prompt shown in Figure \ref{fig:process_prompt}. 
It can be seen that most of the news articles about Alzheimer's Disease in the past year mainly occurred in the United States and Western Europe. 
For the BBC, although it basically only reports Alzheimer's disease news in the UK, the total number of news is not inferior to that of other websites. Similarly, websites in the United States such as NIA also pay more attention to local news, especially in the southeastern states of the United States. For the Alzheimer’s Association, the sources of news reports are relatively scattered all over the world, while the United States and Western Europe still show higher report frequencies than other regions such as 
South America, Africa, Australia, and so on. Finally, for Mayo Clinic, since there is less news from this news source, 
only a few occurrences can be seen on the map. Generally speaking, the distribution of news is worldwide, but it is concentrated in the southeastern United States and Western Europe. These might be because of the select bias of those four news media we use or the well-developed Alzheimer's disease research in these regions. 

Temporal data analysis results can be seen in Figure \ref{fig:temporal}. The numbers of news reports about Alzheimer's disease in each month of the past year (June 2022 to May 2023) are visualized. It can be seen that the overall trend of the number of news reports is declining, from 31 in a single month in June 2022 to 13 in May 2023. It can also be seen that September, October, and November 2022 are the period of high incidences of news reports. The number of news reports in each of the three months exceeded 27, and those in September 2022 reached 32, which was the highest in 2022. This might be because there was news that had a profound impact on AD-related media during this period, resulting in a sudden increase in reports, which deserves special attention from users.

Therefore, it can be seen that \modelname{} can not only extract useful spatiotemporal information from a wide range of news sources but can also use the visualization function to more intuitively display the spatial distribution of the AD-related news and their development through time which might be useful for users. 
We need to emphasize that these spatiotemporal analyses was done by \modelname{} without any human input. 
Thereby \modelname{} improves the efficiency of researchers' work, which AutoGPT  cannot do because it does not design functions of information extraction from web pages.

\subsection{LDA Topic Modeling and Hot Topic Analysis}
\label{sec:hot-topic} 

\begin{figure}[ht!]
	\centering \tiny
\begin{subfigure}[b]{0.48\textwidth}  
		\centering 
		\includegraphics[width=\textwidth]{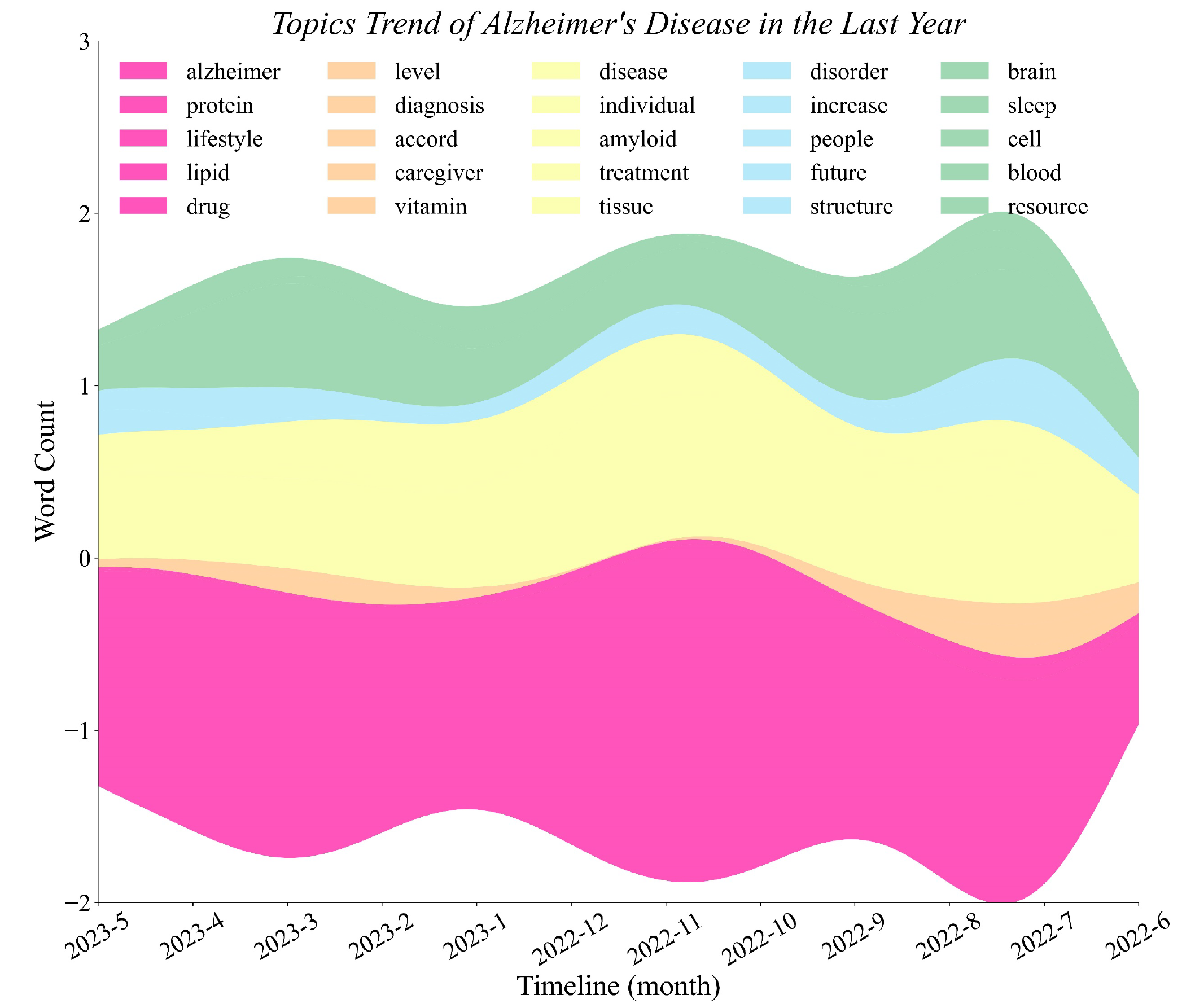}\caption[]{{The word count trend of each topic obtained from the LDA results.
		}}    
		\label{fig:wordCoundTrend}
	\end{subfigure}
	\hfill
	\begin{subfigure}[b]{0.495\textwidth}  
		\centering 
		\includegraphics[width=\textwidth]{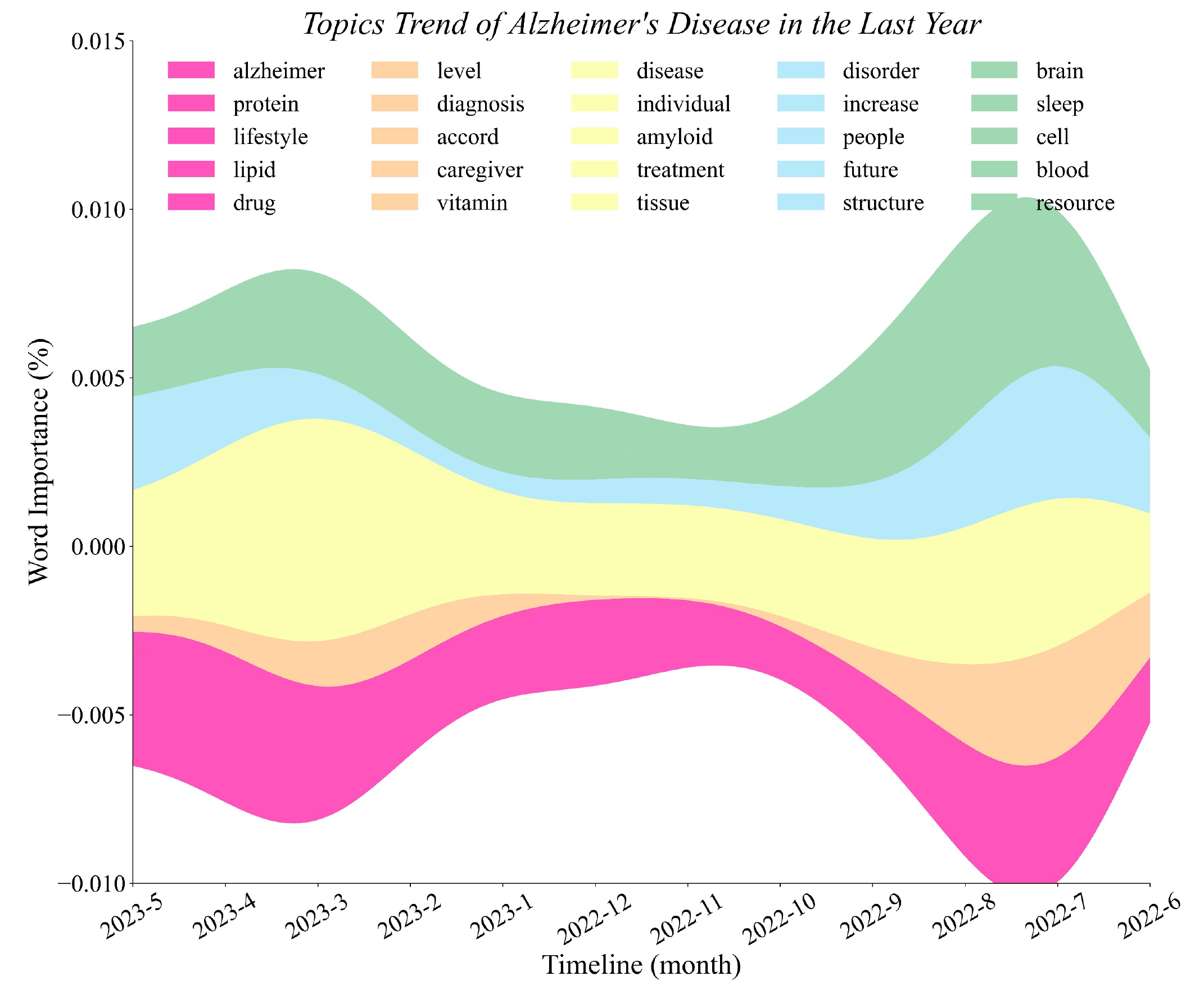}\caption[]{{The word importance trend of each topic obtained from the LDA results.
		}}    
		\label{fig:wordImportanceTrend}
	\end{subfigure}
	\hfill
	\caption{{
	For each Topic, the Streamplot graph displays the
occurrence times and frequency of different keywords in different time periods.}
	} 
	\label{topic_trend}
\end{figure} 
Based on the LDA topic modeling, a hot topic analysis is automatically conducted by \modelname{}. The results can be seen in Figure~\ref{topic_trend}. \modelname{} aggregates the summaries of the news reported in the past year for LDA analysis, and finally got 5 hot topics. It selects the top 5 words with the most occurrences for each of the 5 hot topics and draws streamgraphs according to the number of occurrences and word weights of the words. Please refer to Figure \ref{fig:wordCoundTrend} and \ref{fig:wordImportanceTrend}. In this way, you can see the changes in topic distributions according to time, so as to quickly understand the trend of the research topic. 

It can be found that the keywords of the first hot topic are mainly protein, lipid, and drug, and this type of topic has occupied the largest weight in the past year, which shows that scientists are mostly concerned about seeking reliable drug treatment for Alzheimer's disease. 
The keywords of the topic with the second highest proportion are individual, treatment, amyloid, and tissue. This topic is also about the drug treatment of Alzheimer's disease, but the focus has obviously shifted from the research and development of new drugs to the current personal medication, reflecting the patients' concerns about self-care. The keywords of the third-ranked topic include sleep, brain, blood, cell, etc. This type of news mainly focuses on the causes of Alzheimer’s disease, which is similar to popular science news. It can be seen that journalists have attached great importance to popular science in the past year. 
For the fourth-ranked topic, the keywords are increase, future, disorder, future, etc. 
This topic is mostly related to the future plan or expectation for Alzheimer's disease research. 
The keywords of the last topic are mainly diagnosis, caregiver, vitamin, etc., reflecting the public’s concerns about the diagnosis, care and prevention of Alzheimer’s disease.

Therefore, we can conclude that through hot topic analysis, we can easily get the popular topics in the news 
during June 2022 - May 2023 period by using \modelname's autonomous workflow. 
Users no longer need to read extensively on news, but they can easily use the help of \modelname{} to understand the hot topics of Alzheimer's disease in the past period so that the efficiency of work and research on Alzheimer’s disease is greatly improved. Owing to GPT-4's powerful summarizing ability, in the future, the work of early information collection can be completely handed over to AI. Humans only need to judge and focus on the most critical information returned by AI to quickly understand the development and changes in the public health domain, thus saving time and resources.
 \section{Discussion and Conclusion}

\subsection{Automating Data Analytics}
The success of \modelname{} shows the transformative potential of LLMs in the public health domain. 
By harnessing the advanced linguistic understanding and autonomous operations of \modelname, we were able to streamline the analytical process and conduct comprehensive analyses of extensive news sources related to Alzheimer's Disease (AD). Moreover, \modelname{} has the potential to go beyond the public health domain and be applied in various disciplines.

One of the key advantages of autonomous LLM-based tools such as AutoGPT and \modelname{} is their ability to automate and optimize complex data extraction and analysis tasks, as well as transcending traditional labor-intensive methods. This enables researchers and professionals across different fields to access and engage with large language models, empowering them to conduct sophisticated analyses efficiently, regardless of their technical expertise.

\subsection{Prioritizing Insights and Innovation}
Through the development of \modelname, we conduct a detailed trend analysis, intertopic distance mapping, and identified salient terms relevant to AD. These findings provide valuable insights into the shifting focus and narrative surrounding AD, not only in the domain of public health but also in broader contexts. By quantifying and visualizing the discourse, we gain a nuanced understanding of the prevalent topics, concerns, and perspectives related to AD, facilitating targeted interventions, communication strategies, and decision-making across multiple fields.

The integration of AutoGPT and other autonomous LLM-based tools into research across different disciplines represents a significant advancement. By automating data analysis tasks, researchers can dedicate more time and resources to interpreting the results and deriving actionable insights. This accelerates the research process and enhances the accuracy and reliability of the findings in diverse areas, such as social sciences, economics, technology, and more.

\subsection{Transforming Public Health}
Furthermore, the insights obtained from this research have broader implications beyond public health. The automation capabilities of \modelname{} can revolutionize the field of infodemiology by efficiently analyzing online information trends, tracking the dissemination of information and misinformation, and predicting future trends. This has the potential to inform evidence-based interventions, enhance communication strategies, and combat misinformation across various domains.

While our \modelname{} has made significant strides in utilizing autonomous LLM-based tools for AD analysis in the public health domain, there are still areas for further exploration and improvement.  For example, based on different underlying pathologies, AD-related dementias (ADRD) can be categorized as four major types: prion disease, AD, frontotemporal lobar degeneration (FTLD), and Lewy body diseases (LBD). In practical clinical settings, differentiations among these subtypes of dementias are very challenging, due to both mixed pathologies and clinical symptoms. Our proposed \modelname{} is a general framework and can be easily extended and refined to adapt to other dementias and various brain disorders. Future studies could also focus on expanding the dataset to include a broader range of sources and different languages to capture a more comprehensive understanding of the global discourse on different dementias across different fields. Additionally, exploring the integration of \modelname{} with other data sources, such as social media platforms and electronic records, could provide a more holistic perspective on ADRD conversations and outcomes across multiple disciplines.

\subsection{Ethical Issues related to Autonomous LLM-based Tools}

In the use of autonomous LLM-based tools, several ethical issues arise that warrant careful consideration. First, these models generate output based on their training data, which if biased or discriminatory, could result in outputs that perpetuate such biases \cite{magee2021intersectional,ferrara2023should}. Ethical considerations must therefore include the selection and handling of training data of LLMs to minimize the risk of biased or inappropriate outputs.

In addition, issues of privacy and consent are paramount, particularly when dealing with sensitive data such as health information \cite{liu2023deid}. Even though LLMs do not remember specific inputs or retain personal data, the potential misuse of these tools can lead to leaking private or sensitive information , which raises significant ethical and legal questions.

Moreover, the potential for misuse extends to the propagation of false information or misinformation \cite{liao2023differentiate,hazell2023large}, a concern that is especially salient in the context of public health. LLMs can generate plausible-sounding but factually incorrect or misleading information \cite{liu2023summary,latif2023artificial}, which, if not properly managed, could have severe consequences.

Finally, the democratization of powerful technologies like AutoGPT also raises questions about responsibility and oversight. As these tools become more accessible and widespread, ensuring appropriate use and managing the potential for misuse becomes increasingly challenging.

Addressing these ethical issues is essential for the responsible development and deployment of autonomous LLM-based tools. This includes the development of robust guidelines for data handling, the implementation of safeguards against misuse, the provision of clear user instructions and warnings about potential pitfalls, and ongoing efforts to refine and improve these tools in light of user feedback and societal needs. The goal should be to harness the potential of these technologies while mitigating risks and adverse impacts, striking a balance between technology innovation and ethical responsibility. \section{Conclusion}
In conclusion, this study proposes a transformative autonomous LLM-based tool called \modelname{} which can
facilitate data-driven understanding of complex narratives, not limited to public health but also applicable to various other domains. The initial success of \modelname{} 
has paved the way for future LLM-assisted investigations in global health landscapes and beyond. By leveraging the power of large language models and automation techniques, researchers and professionals can gain valuable insights, inform evidence-based interventions, and drive positive impact across diverse domains.
\label{Discussion} 

\section{Acknowledgments}
This work received support from the National Institutes of Health (NIH) through RO1 grant R01MD013886-05, partial support from grants R01AG075582 and RF1NS128534, as well as support from UGA Interdisciplinary Research Pre-Seed Program -- ``Interdisciplinary Approaches to Alzheimer’s Disease Prevention''. This work is also partially supported by the funding from National Institutes of Health, "Identification of Multi-modal Imaging Biomarkers for Early Prediction of MCI-AD Conversion via Multigraph Representation" (1R03AG078625-01). We would like to express our gratitude to the funding agency for their financial support.

\bibliographystyle{unsrt}

\end{document}